\documentclass{article}

\PassOptionsToPackage{round, authoryear}{natbib}

\usepackage{natbib}
\usepackage{amsmath}
\usepackage{appendix}
\usepackage{makecell}

\usepackage[sglblindworkshop, final]{neurips_2025}

\usepackage{graphicx}

\usepackage[utf8]{inputenc} 
\usepackage[T1]{fontenc}    
\usepackage{hyperref}       
\usepackage{url}            
\usepackage{booktabs}       
\usepackage{amsfonts}       
\usepackage{nicefrac}       
\usepackage{microtype}      
\usepackage{xcolor}         
\usepackage{tabularx}
\usepackage{algorithm}
\usepackage{algpseudocode}
\usepackage{subcaption}
\usepackage{float}
\title{DynaStride: Dynamic Stride Windowing with MMCoT for Multi-Scene Captioning}

\author{
Eddison Pham\\
Algoverse AI Research\\
\texttt{eddison.pham@mail.utoronto.ca}
\And
Prisha Priyadarshini\\
Algoverse AI Research\\
\texttt{pp1155@scarletmail.rutgers.edu}
\And
Adrian Maliackel\\
Algoverse AI Research\\
\texttt{amaliack@purdue.edu}
\And
Kanishk Bandi\\
Algoverse AI Research\\
\texttt{kanishk.bandi@mail.utoronto.ca}
\And
Cristian Meo\\
Algoverse AI Research\\
LatentWorlds AI\\
Delft University of Technology\\
\texttt{c.meo@tudelft.nl}
\And
Kevin Zhu\\
Algoverse AI Research\\
\texttt{kevin@algoverse.us}
}

\workshoptitle{NeurIPS 2025 Workshop on Vision-Language Intelligence}
\begin{document}

\maketitle
\begin{abstract}
Scene-level captioning in instructional videos can enhance learning by requiring an understanding of both visual cues and temporal structure. By aligning visual cues with textual guidance, this understanding supports procedural learning and multimodal reasoning, providing a richer context for skill acquisition. However, captions that fail to capture this structure may lack coherence and quality, which can create confusion and ultimately undermine the video's educational intent. To address this gap, we introduce DynaStride, a pipeline to generate coherent, scene-level captions without requiring manual scene segmentation. Using the YouCookII dataset’s scene annotations, DynaStride performs adaptive frame sampling and multimodal windowing to capture key transitions within each scene. It then employs a multimodal chain-of-thought process to produce multiple action-object pairs, which are refined and fused using a dynamic stride window selection algorithm that adaptively balances temporal context and redundancy. The final scene-level caption thus integrates visual semantics and temporal reasoning in a single instructional caption. Empirical evaluations against strong baselines, including VLLaMA3 and GPT-4o, demonstrate consistent gains on both N-gram-based metrics (BLEU, METEOR) and semantic similarity measures (BERTScore, CLIPScore). Qualitative analyses further show that DynaStride produces captions that are more temporally coherent and informative, suggesting a promising direction for improving AI-powered instructional content generation. Code for this work is available on GitHub.\footnotemark[1]
\end{abstract}

\footnotetext[1]{\url{https://github.com/eddisonpham/DynaStride}}

\section{Introduction}

Instructional videos are widely used to teach complex tasks by providing step-by-step guidance that can be modeled through deep learning \citep{narasimhan2023learning}. These videos typically represent a sequence of steps, each with distinct scenes that highlight specific stages of the task. Recent work shows that instructional video corpora often contain rich multimodal signals which can be exploited to segment and caption procedural steps, but also introduce challenges of alignment, redundancy, and temporal coherence \citep{shi2019dense}. Generating accurate scene-level captions is crucial for accessibility, content summarization, and video search, enabling learners to quickly grasp sequences of actions and the underlying concepts \citep{yousif2023survey}. High-quality captions also improve accessibility for visually impaired or differently abled learners and allow educators to efficiently index and summarize content \citep{gernsbacher2015video}. In addition, instructional videos have been shown to significantly improve learning outcomes, with a meta-analysis revealing a strong effect size of 0.688 when video clips are thoughtfully integrated into lessons \citep{lin2023meta}. As such, advances in scene-level caption generation have significant potential to enhance personalized learning experiences, enable adaptive recommendations, and support the delivery of scalable educational content. 

Artificial Intelligence (AI) and Machine Learning (ML) have increasingly transformed educational technologies, allowing adaptive learning, personalized feedback, and data-driven insights into student performance \citep{elstad2024aieducationrationaleprinciples, moralesnavarro2024unpackingapproacheslearningteaching}. In the context of instructional videos, AI-powered video analysis and automated captioning systems can improve accessibility, facilitate content indexing, and support personalized learning experiences \citep{shvetsova2024howtocaptionpromptingllmstransform}. Deep learning models, including vision-language architectures, allow for automated understanding of complex visual and temporal sequences, providing learners and educators with actionable insights from video content \citep{chu2025finegrained, vid2seq2023}. Together, these AI and ML advancements promise to make instructional video-based learning more accessible, engaging, and tailored to individual learner needs.

Recent research has highlighted the importance of scene-level captions in educational videos. For instance, \citep{malakul2023auto} demonstrated that automatically generated subtitles can significantly enhance both learning comprehension and learner satisfaction compared to videos without subtitles. Advances in video captioning, such as leveraging scene graphs for fine-grained captioning, have further shown promise in producing captions that are more accurate and contextually aware, especially for longer videos \citep{chu2025finegrained}. Additionally, \citep{alabsi2020subtitles} found that using apps to add subtitles to educational videos can improve students' listening comprehension, indicating that the way subtitles are presented can meaningfully affect learning outcomes. These studies demonstrate that high-quality, scene-level captions can significantly improve how educational videos support learning and comprehension.

Despite significant progress in scene-level captioning, current methods face notable challenges in accurately representing instructional videos. Many approaches attempt to generate captions for every frame or segment, often producing dense descriptions that include redundant or irrelevant information \citep{auroracap2024, vid2seq2023}. In contrast, shorter caption methods frequently miss important details about actions, objects, or their temporal relationships, resulting in incomplete or ambiguous descriptions \citep{tang2025caption}. Furthermore, some systems rely heavily on localized inference tools, which can limit scalability, reproducibility, and the ability to maintain consistent quality in long or complex videos \citep{auroracap2024}. Ultimately, these challenges point to the need for smart, context-sensitive captioning that captures key actions and objects, reduces repetition, and preserves the correct sequence of events.

In this work, we introduce DynaStride, a scalable framework for scene-level captioning on educational videos, evaluated on the YouCookII dataset \citep{zhou2018youcookii}, which leverages existing scene boundaries and human-labeled ground-truth captions for cooking videos (an extended overview of the dataset can be found in Appendix \ref{appendix: dataset}). By leveraging existing scene boundaries and human-labeled ground-truth captions, DynaStride avoids the need for custom segmentation. This framework adaptively skips redundant frames, applies multimodal chain-of-thought (MMCoT) prompting \citep{zhang2025cmmcot} to generate multiple detailed subcaptions, and finally aggregates them into concise, scene-level descriptions. When evaluating DynaStride against state-of-the-art vision-language models, including GPT-4o \citep{openai2024gpt4o} and VideoLLaMA-3 \citep{llama3_2024}, our framework demonstrates improvements in both N-gram and semantic-based metrics.

\section{Related Works}
\label{gen_inst}

\subsection{Video Captioning} 

Video captioning research has focused on improving temporal understanding and long-form video comprehension using encoder-decoder and transformer-based architectures that capture dense, action-oriented representations \cite{Xie2024, Wu2025, Muksimova2025}. Hierarchical approaches, such as AuroraCap and ShareGPT4Video \cite{chai2025auroracap, chen2024sharegpt4video}, decompose the task into stages including feature extraction, event localization, and caption generation across temporal scales , while multi-level frameworks generate captions at the frame, clip, and video levels \cite{wei2025longcaptioning}. Other methods enhance temporal reasoning with mechanisms like Timestamp Injection Mechanism (TIM) and Temporal-Aware Similarity Sampling (TASS) \cite{date}, or use causal-temporal narrative structures for coherent descriptions \cite{narrativebridge}. Reinforcement learning and intention-oriented controllable captioning improve fine-grained or targeted descriptions \cite{videocap_r1,intentvcnet}, and spatial-temporal and multimodal modules like STC enhance video-language modeling \cite{videollama2}. Building on these strategies, DynaStride aggregates frame and scene-level captions, combining temporal reasoning with long-form video understanding.

\subsection{Chain-of-Thought Reasoning}
\citet{wei2022chain} introduced Chain-of-Thought (CoT) prompting, a technique that enhances the reasoning capabilities of large language models by guiding them to generate intermediate reasoning steps before producing a final answer. This approach has been particularly effective for tasks requiring multi-step reasoning, such as arithmetic, commonsense, and symbolic reasoning, by encouraging models to mimic human-like thought processes. Building on this idea, \citet{cmmcot} proposed the Complex Multi-Modal Chain-of-Thought (CMMCoT) framework, which extends CoT reasoning to complex multi-image comprehension tasks. CMMCoT employs interleaved multimodal reasoning chains, where critical visual region tokens extracted from intermediate reasoning steps serve as supervisory signals, improving cross-modal understanding and interpretability. Additionally, it incorporates test-time memory augmentation through the Retrieval-based Image Feature Reasoning Enhancement Module (RIFREM), which stores key and value pairs of image features from each decoder layer in a memory bank. This design enables efficient cross-attention computations and allows the model to dynamically recall relevant visual information during reasoning, effectively expanding its reasoning capacity without increasing the number of parameters. Collectively, these studies highlight the growing importance of structured reasoning and dynamic memory mechanisms for advancing multimodal reasoning across video frames.

\section{Methodology}
To achieve our goal of generating high-quality instructional captions for multiple scenes, we propose the following multi-scene to text pipeline: 
(1) frame sampling and windowing, (2) MMCoT subcaption generation, (3) a dynamic stride window selection algorithm, and (4) subcaption aggregation.

\begin{figure}[ht]
    \centering
    \vspace{-2.5mm} 
    \includegraphics[width=\linewidth,clip,trim=6pt 6pt 6pt 6pt]{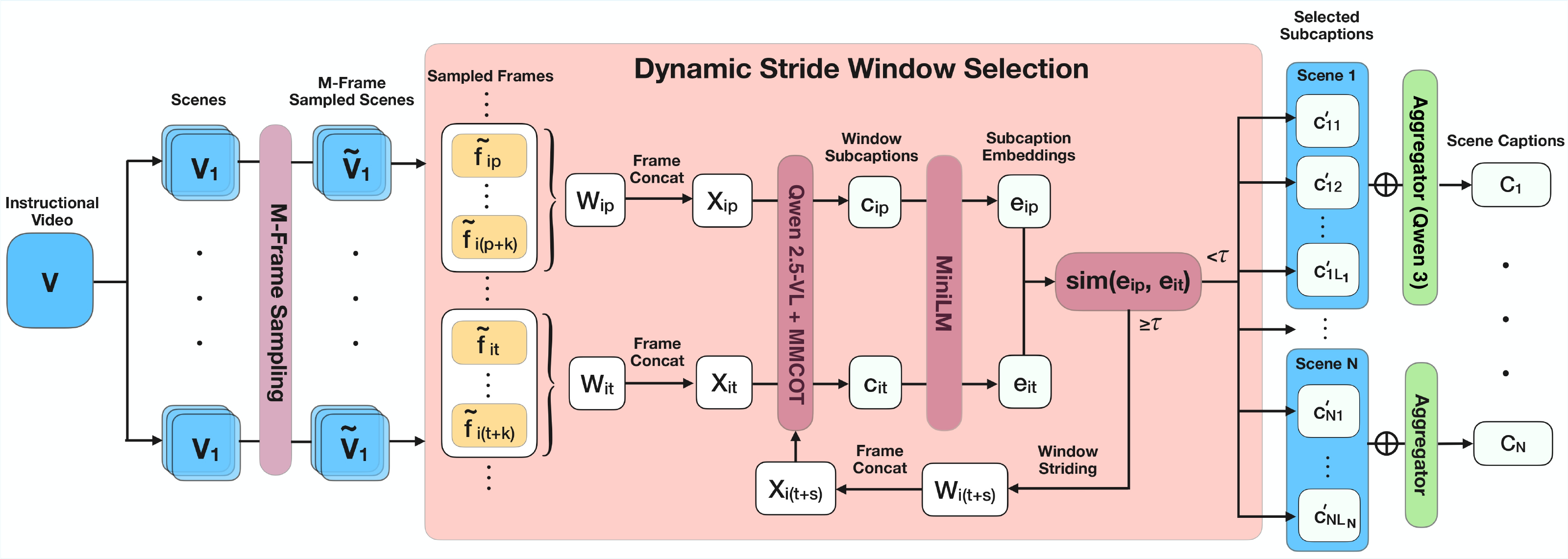}
    \vspace{-3mm}
    \caption{\footnotesize Illustration of the multi-scene-to-text captioning pipeline depicting four stages: frame sampling, MMCoT, dynamic stride, and aggregation.}
    \label{fig: model-architecture}
    \vspace{-3mm}
\end{figure}

The goal of this pipeline is to reduce content redundancy in scenes by eliminating semantically similar windows and retaining only the most informative ones. This allows MMCoT to more effectively track actions over time and reason about the objects involved. Pretrained models were used without fine-tuning to ensure the reproducibility of our pipeline. Figure \ref{fig: model-architecture} illustrates the proposed methodology.

\subsection{Frame Sampling and Windowing}  
To efficiently process long video scenes, we first define the $i$-th scene of a video $V$ as an ordered sequence of frames:
\[
V_i = [f_{i1}, f_{i2}, \dots, f_{iN_i}], \quad f_{ik} \in \mathbb{R}^{H \times W \times C}, \quad 1 \le k \le N_i
\]
where $N_i$ is the total number of frames in the scene, and $H, W, C$ denote the height, width, and number of color channels of each frame, respectively.  

Directly processing all frames in a long scene can be computationally expensive and often redundant, as consecutive frames typically contain highly similar visual information. To address this, we subsample the scene by selecting every $M$-th frame, producing a more compact subsequence:
\[
\tilde{V}_i = [\tilde{f}_{i1}, \tilde{f}_{i2}, \dots, \tilde{f}_{i\lfloor N_i/M \rfloor}]
\]

Next, to capture localized temporal dynamics while keeping computations manageable, we define a sliding window $W_{it}$ starting at index $t$ as a consecutive block of $K$ frames from the subsampled sequence:
\[
W_{it} = [\tilde{f}_{it}, \tilde{f}_{i(t+1)}, \dots, \tilde{f}_{i(t+K)}].
\]
This windowing strategy allows our model to focus on short-term temporal patterns within each scene, facilitating more precise feature extraction and downstream caption generation, while avoiding unnecessary processing of redundant frames.
\subsection{MMCoT Subcaption Generation}  
For each window $W_{it}$, we first concatenate its frames horizontally to form a wide image:
\[
X_{it} = \text{Concat}(W_{it}) \in \mathbb{R}^{H \times (|W_{it}| \cdot W) \times C},
\]

which allows a vision-language model to perceive temporal context across multiple frames in a single input, rather than processing each frame independently.  

We then employ \texttt{Qwen2.5-VL-Instruct} ($\mathcal{Q}$), an instruction tuned image-to-text model, to generate a \emph{subcaption} describing both the observed action and associated objects:
\[
c_{it} = \mathcal{Q}(X_{it}) = \text{"$a_{it} \,|\, o_{it}$"},
\]  
where $a_{it}$ represents the action description, $o_{it}$ is the set of objects present, and the delimiter ``$|$'' separates these components to facilitate downstream aggregation. This multimodal chain-of-thought approach encourages the model to reason over both temporal and semantic content simultaneously.

To reduce redundancy and improve efficiency, we adaptively select windows using a stride parameter $s$. Let $p$ denote the index of the most recently retained window with subcaption $c_{ip}$. For a candidate window $W_{it}$, we first compute its embedding using \texttt{MiniLM} ($\mathcal{M}$) \cite{wang2020minilm}:
\[
e_{it} = \mathcal{M}(c_{it}), \quad e_{ip} = \mathcal{M}(c_{ip}),
\]  
and measure similarity via cosine similarity: $\text{sim}(e_{it}, e_{ip})$.  

If $\text{sim}(e_{it}, e_{ip}) \ge \tau$, the candidate window is skipped, and the stride is increased according to
\[
s \gets \min(\alpha \cdot s, s_{\max}),
\]  
\subsection{Subcaption Aggregation}  
The retained subcaptions $\{c_{i1}', \dots, c_{iL_i}'\}$ are concatenated and fed into \texttt{Qwen3} ($\mathcal{A}$) \cite{qwen3_2024}, to produce a final instructional caption for the scene:
\[
C_i = \mathcal{A}(c_{i1}' \oplus \cdots \oplus c_{iL_i}'),
\]  
where $\oplus$ denotes stirng concatenation, with each string separated by the newline character \texttt{\textbackslash n}.
 This approach ensures that the resulting caption captures the temporal evolution of actions and identifies objects most relevant to each change.  

Implementation details, including parameter choices for $M$, $K$, $\alpha$, $s_{\text{base}}$, and $\tau$, are provided in Appendix \ref{appendix: implementation details}. 

\section{Experiments}
\label{others}

In this section, we conducted empirical experiments to understand the performance of DynaStride and its
potential limitations by exploring the following questions: \begin{itemize}
    \item \textbf{RQ 1.} To what extent does our method improve the coherence and meaningfulness of the inferred scene captions? 
    \item \textbf{RQ 2.} How does the frame sampling and aggregator impact overall caption quality?
\end{itemize}

\subsection{Experimental and Ablation Setup}
We evaluated DynaStride against two SoTA VLMs, GPT-4o \citep{openai2024gpt4o} and VideoLLaMA-3 \citep{llama3_2024}, using the same prompt for both baselines and the aggregator model to ensure fair comparison. Experiments and ablations were conducted on 210 YouCookII videos \citep{zhou2018youcookii} with three seeds, reporting mean and standard deviation for all metrics. For computing, we used 30× RTX A6000 GPUs and 800 GB of storage. Performance was measured across multiple captioning quality metrics covering N-gram overlap, semantic alignment, TF-IDF consensus, and embedding similarity: BLEU-4 \citep{papineni2002bleu}, METEOR \citep{banerjee2005meteor}, CIDEr \citep{vedantam2015cider}, BERTScore \citep{zhang2020bertscore} (Precision, Recall, F1), SBERT Similarity \citep{reimers2019sentencebertsentenceembeddingsusing}, and Temporal Analysis \citep{sun2021nsp, ilharco2019general} (see Appendix \ref{appendix: metric details} for details). We also investigated the impact of frame sampling rates and aggregator choice on caption quality, comparing sampling rates (5, 20, 40) and aggregators including Phi \citep{abdin2024phi3} and Mistral \citep{jiang2023mistral}. Direct caption examples and graphs of frame sampling rates versus performance metrics are in Appendices \ref{appendix: caption comparison} and \ref{appendix: frame sampling vs metrics}.

\subsection{Experiment Results}
\vspace{-0.25cm}
\begin{table}[ht]
\centering
\footnotesize
\setlength{\tabcolsep}{5pt} 
\begin{tabularx}{\textwidth}{>{\hsize=5\hsize\raggedright}X *{3}{c} *{3}{c} c}
\toprule
Method & 
\multicolumn{3}{c}{N-gram} & 
\multicolumn{3}{c}{BERT} & 
SBERT \\
\cmidrule(lr){2-4} \cmidrule(lr){5-7} \cmidrule(lr){8-8}
 & B@4($\uparrow$) & METEOR($\uparrow$) & CIDEr($\uparrow$) & Prec($\uparrow$) & Recall($\uparrow$) & F1($\uparrow$) & (Sim$\uparrow$) \\
\midrule
GPT-4o & \textbf{4.73}(0.63) & \textbf{28.47}(1.37) & 0.48(0.06) & 0.19(0.01) & \textbf{0.29}(0.01) & 0.23(0.01) & 0.60(0.01) \\
VLLaMA-3 &  4.10(0.32) & 22.71(0.63) &  0.49(0.02) & 0.19(0.01) & 0.22(0.00) & 0.21(0.01) & 0.58(0.00) \\
DynaStride & 4.18(0.07) & 24.31(0.10) & \textbf{0.56}(0.00) & \textbf{0.25}(0.00) & 0.26(0.00) & \textbf{0.27}(0.00) & \textbf{0.61}(0.00) \\
\bottomrule
\end{tabularx}
\vspace{0.1cm}
\caption{\footnotesize Scene captioning results on YouCook2 validation set, comparing GPT-4o, LLaMA-3, and our method.}
\label{table:youcook2-all-metrics}
\end{table}

\textbf{DynaStride achieves the highest CIDEr score and semantic metrics compared to the baselines.}
As shown in Table \ref{table:youcook2-all-metrics}, DynaStride consistently outperforms VLLaMA-3 across all metrics, with notable gains in CIDEr, precision, F1, and SBERT similarity. Compared to GPT-4o, DynaStride achieves higher scores on CIDEr, BERT precision, BERT F1, and SBERT similarity, indicating that our captions are not only more relevant but also semantically closer to the ground-truth references. These results highlight that our method produces captions that are both more accurate and contextually meaningful. Further results on temporal performance can be found in Appendix \ref{appendix: temporal metrics}.

\subsection{Ablation Results}

\begin{table}[ht]
\centering
\footnotesize
\setlength{\tabcolsep}{5pt}

\begin{tabularx}{\textwidth}{>{\hsize=5\hsize\raggedright}X *{3}{c} *{3}{c} c}
\toprule
Aggregator & 
\multicolumn{3}{c}{N-gram} & 
\multicolumn{3}{c}{BERT} & 
SBERT \\
\cmidrule(lr){2-4} \cmidrule(lr){5-7} \cmidrule(lr){8-8}
 & B@4($\uparrow$) & METEOR($\uparrow$) & CIDEr($\uparrow$) & Prec($\uparrow$) & Recall($\uparrow$) & F1 ($\uparrow$) & Sim($\uparrow$) \\
\midrule
Phi & 2.78(0.24) & 22.8(0.38) & 0.37(0.02) & 0.17(0.01) & 0.25(0.00) & 0.21(0.01) & 0.59(0.00) \\
Mistral & 3.36(0.08) & 19.49(0.12) & 0.51(0.01) & \textbf{0.27}(0.00) & 0.23(0.00) & 0.25(0.00) & 0.6(0.00) \\
Qwen3 & \textbf{4.18}(0.07) & \textbf{24.31}(0.10) & \textbf{0.56}(0.00) & 0.25(0.00) & \textbf{0.26}(0.00) & \textbf{0.27}(0.00) & \textbf{0.61}(0.00) \\
\bottomrule
\end{tabularx}
\vspace{0.1cm}
\caption{\footnotesize Comparison of metrics using different aggregator models (Phi, Mistral, Qwen3).}
\label{table:aggregator-models}
\end{table}

\begin{table}[ht]
\centering
\footnotesize
\setlength{\tabcolsep}{5pt}

\begin{tabularx}{\textwidth}{>{\hsize=5.5\hsize\raggedright}X *{3}{c} *{3}{c} c}
\toprule
Model / Rates & 
\multicolumn{3}{c}{N-gram} & 
\multicolumn{3}{c}{BERT} & 
SBERT \\
\cmidrule(lr){2-4} \cmidrule(lr){5-7} \cmidrule(lr){8-8}
 & B@4($\uparrow$) & METEOR($\uparrow$) & CIDEr($\uparrow$) & Prec($\uparrow$) & Recall($\uparrow$) & F1($\uparrow$) & Sim($\uparrow$) \\
\midrule
\textbf{GPT-4o} & & & & & & & \\
\quad 5 Frames & 4.31(0.15) & 27.58(0.47) & 0.45(0.02) & 0.18(0.00) & 0.28(0.00) & 0.23(0.00) & 0.59(0.00) \\
\quad 20 Frames & 4.69(0.19) & 27.91(0.28) & 0.49(0.02) & 0.19(0.01) & 0.29(0.00) & 0.24(0.00) & 0.60(0.00) \\
\quad 40 Frames & 4.52(0.11) & \textbf{28.07}(0.03) & 0.48(0.00) & 0.19(0.00) & \textbf{0.29}(0.00) & 0.24(0.00) & 0.60(0.00) \\
\textbf{VLLaMA-3} & & & & & & & \\
\quad 5 Frames & 3.60(0.45) & 22.48(0.17) & 0.45(0.05) & 0.18(0.00) & 0.22(0.00) & 0.19(0.00) & 0.57(0.00) \\
\quad 20 Frames & 4.32(0.29) & 22.28(0.27) & 0.52(0.02) & 0.22(0.00) & 0.22(0.00) & 0.22(0.00) & 0.58(0.00) \\
\quad 40 Frames & 4.79(0.05) & 21.9(0.08) & 0.56(0.01) & \textbf{0.27}(0.00) & 0.21(0.00) & 0.24(0.00) & 0.58(0.00) \\
\textbf{DynaStride} & & & & & & & \\
\quad 5 Frames & 3.89(0.12) & 23.38(0.14) & 0.52(0.01) & 0.24(0.00) & 0.25(0.00) & 0.24(0.00) & 0.60(0.00) \\
\quad 20 Frames & 4.48(0.03) & 24.82(0.1) & 0.58(0.00) & 0.24(0.00) & 0.26(0.00) & 0.25(0.00) & 0.61(0.00) \\
\quad 40 Frames & \textbf{4.91}(0.03) & 26.36(0.18) & \textbf{0.61}(0.00) & 0.25(0.00) & 0.28(0.00) & \textbf{0.27}(0.00) & \textbf{0.63}(0.00) \\
\bottomrule
\end{tabularx}

\vspace{0.1cm}
\caption{\footnotesize Comparison of different frame sampling rates (5, 20, and 40 frames).}
\label{table:frame-sampling-rates}
\end{table}

\textbf{Sparse sampling boosts caption quality, and aggregator choice impacts accuracy and consistency.}
Based on the metrics in Table 4, higher frame sampling generally improves caption quality across all models, though the effect varies by method. For GPT-4o, increasing from 5 to 20 frames yields consistent gains in B@4, METEOR, and CIDEr, plateauing between 20 and 40 frames. VLLaMA-3 shows steady improvement up to 40 frames, achieving the highest CIDEr and F1 scores, although SBERT similarity remains lower than that of GPT-4o. DynaStride benefits most from sparser sampling, with 20 and 40 frames producing consistent gains across all metrics, reaching the highest overall scores at 40 frames; particularly CIDEr (0.61), F1 (0.27), and SBERT similarity (0.63). Separately, Table 3 shows that aggregator choice has a strong impact on N-gram metrics. Qwen3 dominates overall performance in all but BERT Precision, and has the lowest standard deviation across all metrics. Furthermore, we found that Phi's standard deviation metrics were more than three times higher than Qwen3, indicating that it had less consistent caption generations. These results indicate that both dense frame sampling and careful aggregator selection are crucial for not only producing coherent and semantically meaningful captions, but also consistency in caption generation.

\section{Limitations}

Despite its promising performance, DynaStride has several limitations that are important to consider. Its reliance on pretrained models such as Qwen2.5-VL and Qwen3 raises concerns about generalization beyond the YouCookII domain, as these models may not fully capture domain-specific nuances or procedural variations in other instructional contexts. Similarly, the YouCookII dataset, while widely used, is relatively small and may not reflect the full diversity of instructional tasks, which could limit the applicability of the generated captions to other domains or more complex procedural workflows.

Additionally, the current architecture does not explicitly model cross-modal interactions over long temporal spans, which may reduce its ability to capture complex dependencies between actions, objects, and instructional narration. While dynamic frame sampling improves computational efficiency, it may overlook subtle or rapid actions, potentially producing incomplete, ambiguous, or temporally inconsistent subcaptions. Although the dynamic stride algorithm reduces redundant frames, it may not completely mitigate this issue during subcaption aggregation, potentially impacting the coherence, clarity, and instructional value of the final scene-level captions.

Moreover, standard automatic evaluation metrics, such as N-gram overlap or semantic similarity scores, fail to fully capture human judgments of instructional relevance, temporal accuracy, or educational usefulness, which are crucial for learning outcomes. Finally, DynaStride does not currently incorporate mechanisms for domain adaptation or active feedback, limiting its ability to continuously improve in new instructional settings or personalize captions for diverse learner populations.

\section{Conclusion}

This work introduces DynaStride, a pipeline that leverages YouCookII annotations for scene-level captioning in instructional videos. By applying frame sampling to focus on the most informative portions of each scene, generating MMCoT action-object subcation pairs with Qwen2.5-VL, filtering redundant pairs with the dynamic windowing algorithm, and aggregating them into coherent scene-level captions with Qwen3, DynaStride efficiently captures key instructional content. Thus, DynaStride is able to support educational tasks such as lesson indexing, content retrieval, comprehension, and accessibility, aligning with broader efforts to enhance learning through AI-powered instructional content generation.

DynaStride offers several practical benefits. Dynamic frame sampling reduces computational overhead while retaining essential visual information, allowing faster processing without major sacrifices in caption quality. Aggregation of subcaptions ensures that scene-level captions remain coherent, informative, and temporally consistent, effectively conveying sequences of actions critical for understanding procedural content. Quantitative evaluation shows that DynaStride outperforms baseline models across N-gram and semantic metrics, including CIDEr, precision, F1, and SBERT similarity, indicating strong performance in both lexical accuracy and semantic understanding.

Despite these strengths, several limitations remain, including reliance on pretrained models, limited domain diversity in YouCookII, and challenges in modeling long-range cross-modal dependencies. Addressing these limitations represents a promising direction for future work. Extending the pipeline to handle raw, unsegmented videos will require robust scene boundary detection, potentially through temporal action detection models or weakly supervised segmentation techniques. Adaptive frame sampling informed by visual saliency, action dynamics, or instructional importance could further optimize coverage while maintaining efficiency. Improved caption aggregation that models procedural coherence and temporal alignment could enhance instructional clarity. Finally, incorporating human evaluations and expanding to diverse instructional domains would ensure that captions are both practically useful and educationally impactful.
\bibliographystyle{plainnat}
\bibliography{references}      
\newpage

\appendix
\section{Dataset}
\label{appendix: dataset}
The YouCookII dataset consists of 2,000 cooking videos annotated with scene boundaries and corresponding procedural descriptions of the scenes. As the test set captions were not provided, we uniformly sampled $210$ videos from the validation set to be used on all experiments.\\

Efficient preprocessing of the YouCookII dataset is critical for enabling downstream scene captioning tasks. We developed a robust pipeline that converts raw videos and annotations into scene-segmented, frame-level data for model training and evaluation.\\

Starting from the official YouCookII dataset archive, which contains raw videos, annotations, and split lists, we organized the dataset by split (train, validation, test) and recipe type to streamline batch processing. We replaced the original downloader’s youtube-dl with yt-dlp for improved reliability, handling restricted videos via manual cookie export and authentication to ensure successful downloads in .mp4 format with integrity verification.\\

For each split, a preprocessing script (preprocessor.py) parsed JSON annotations to extract scene boundary metadata (startFrame, endFrame) into .pkl files. Videos were segmented accordingly, and frames within each scene were sampled and resized to 384×384 pixels to standardize inputs. Scene frames were saved in uniquely named directories (<video\textunderscore id>\textunderscore <segment\textunderscore num>) under respective split folders.

\section{Implementation Details}
\label{appendix: implementation details}

\subsection{Models and Access}

All the models used for this study can be accessed through HuggingFace.

\textbf{MMCoT Subcaptioning:} 

Window-level captions are generated using the \texttt{Qwen2.5-VL-7B-Instruct} model, which produces action-object subcaptions for each video window.

\textbf{Aggregation Models:}  

For aggregating subcaptions into a single instructional caption, the following models were considered:
\begin{itemize}
    \item \texttt{Mistral-7B-Instruct-v0.2}  
    \item \texttt{Qwen3-4B-Instruct-2507}  
    \item \texttt{Phi-3-mini-4k-instruct}  
\end{itemize}
All of these aggregation models are freely available on HuggingFace.

\subsection{MMCoT Prompting}

The \texttt{Qwen2.5-VL-7B-Instruct} model is used to reason about the temporal progression of actions within a window of frames. To reduce object hallucinations and ensure concise outputs, the model is prompted to produce the following:
\begin{enumerate}
    \item The action being performed in the sequence.  
    \item A list of objects involved.
\end{enumerate}
These outputs are separated by a ``|'' character, without showing any internal reasoning. The model internally reasons over the sequence, producing short, clear descriptions that capture the full temporal progression. \texttt{<CONCLUSION></CONCLUSION>} tokens are used to surround our answer for downstream pipelines to automatically identify and extract outputs.
\newpage

\textbf{MMCoT Prompt:}
\begin{quote}
USER: <image> <image> <image> ... \\
These images show a sequence of events from left to right. \\
Task: 
\begin{enumerate}
    \item Carefully reason about the sequence of actions in the images (internally).
    \item Produce exactly two outputs separated by a ``|'': 
        \begin{itemize}
            \item Output 1: Description of the exact action being performed throughout the sequence.
            \item Output 2: List of objects involved in the sequence.
        \end{itemize}
    \item Do NOT show internal reasoning or extra captions—only the final outputs.
    \item Keep it short, clear, and concise.
\end{enumerate}
Output format: \texttt{<CONCLUSION>Action | Objects</CONCLUSION>} \\
Focus on the full temporal progression of the sequence.
\end{quote}

\subsection{Subcaption Aggregation}

After generating window-level captions, subcaptions are aggregated into a single concise instruction using \texttt{Qwen3-4B-Instruct}. The aggregation prompt enforces:
\begin{itemize}
    \item \textbf{Chronological consistency:} Captions are processed in temporal order.
    \item \textbf{Imperative tone:} The final output is a short instruction sentence.
    \item \textbf{Conciseness:} Only one sentence is produced, enclosed in \texttt{<ANSWER>} and \texttt{</ANSWER>}.
    \item \textbf{No extra reasoning:} Intermediate steps or explanations are omitted.
\end{itemize}

\textbf{Aggregator Prompt:}
\begin{quote}
You are given multiple captions from a short instructional clip, in chronological order. \\
Write ONE concise sentence that is short and instructional. \\
Use an imperative tone, as if giving instructions for performing a task. \\
Your response MUST be enclosed between <ANSWER> and </ANSWER>, containing ONLY the final instruction sentence. \\[1ex]
Captions: \\
1. <first caption> \\
2. <second caption> \\
... \\
L. <last caption> \\[1ex]
Output:
\end{quote}

\subsection{Dynamic Stride Window Selection}

\textbf{Dynamic Stride Parameters.}  
The following parameters were used for this algorithm: 
\begin{itemize}
    \item Window size: $K = 10$
    \item Base stride: $s_{\text{base}} = 10$
    \item Initial stride: $s = 10$
    \item Maximum stride: $s_{\max} = 3 \cdot s_{\text{base}}$
    \item Stride scaling factor: $\alpha = 1.5$
    \item Similarity threshold: $\tau = 0.5$
\end{itemize}
Subcaption embeddings are computed using \texttt{all-MiniLM-L6-v2}.
\newpage
\textbf{Cosine Similarity.}  
For two subcaptions $c_i$ and $c_j$, let $e_i, e_j$ be their embeddings from \texttt{all-MiniLM-L6-v2}. Their similarity is computed as:
\[
\text{sim}(e_i, e_j) = \frac{\langle e_i, e_j \rangle}{\|e_i\| \, \|e_j\|},
\]
where $\langle \cdot, \cdot \rangle$ denotes the dot product and $\|\cdot\|$ is the Euclidean norm.

\textbf{Pseudocode for Dynamic Stride Window Selection.}

\begin{algorithm}[H]
\caption{Dynamic Stride Window Selection}
\begin{algorithmic}[1]
\Function{DynamicStrideWindowSelection}{$V_i$, $K$, $M$, $s_{\text{base}}$, $s_{\max}$, $\alpha$, $\tau$}
    \State $\tilde{V}_i \gets$ subsample every $M$-th frame from $V_i$
    \State $\mathrm{captions} \gets [\,]$
    \State $e_{ip} \gets \text{None}$ \Comment{last retained embedding}
    \State $s \gets s_{\text{base}}$
    \State $t \gets 0$
    \While{$t < |\tilde{V}_i|$}
        \State $W_{it} \gets \tilde{V}_i[t : t+K]$ \Comment{current window}
        \State $X_{it} \gets \operatorname{Concat}(W_{it})$
        \State $c_{it} \gets \mathcal{Q}(X_{it})$ \Comment{subcaption for the window}
        \State $e_{it} \gets \mathcal{M}(c_{it})$ \Comment{embedding of the subcaption}
        \If{$\operatorname{sim}(e_{it}, e_{ip}) \ge \tau$}
            \State $s \gets \min(\alpha \cdot s, s_{\max})$ \Comment{skip window, increase stride}
        \Else
            \State $\mathrm{captions}$.append($c_{it}$)
            \State $e_{ip} \gets e_{it}$ \Comment{update most recent embedding}
            \State $s \gets s_{\text{base}}$ \Comment{reset stride}
        \EndIf
        \State $t \gets t + \max(1, s)$
    \EndWhile
    \State \Return $\mathrm{captions}$
\EndFunction
\end{algorithmic}
\end{algorithm}

\section{Evaluation Metrics Details}
\label{appendix: metric details}

We evaluated generated captions against reference captions using several standard metrics. For BLEU-4 and METEOR, we report the models' ability to generate captions that align with reference N-grams and semantic content, respectively. CIDEr evaluates consensus with reference captions using TF-IDF weighting, while BERTScore and SBERT similarity scores measures semantic similarity at the embedding level, with separate values for precision (PBERT), recall (RBERT), and F1 (FBERT). We also did temporal analysis on metrics: Align$_\text{DTW}$, Contradict$_\text{NLI}$, NSP=True, NSP=Shuffled, and NSP=Delta. 

\textbf{BLEU-4 \citep{papineni2002bleu}.}  
The BLEU-4 score measures the overlap of 4-grams between a generated caption and reference captions. For a candidate caption $c$ and a set of reference captions $\{r_1, \dots, r_m\}$, the BLEU-4 score is defined as
\[
\text{BLEU-4}(c, \{r_k\}) = \text{BP} \cdot \exp\Big( \frac{1}{4} \sum_{n=1}^{4} \log p_n \Big),
\]
where $p_n$ is the precision of N-grams of size $n$, and BP is the brevity penalty:
\[
\text{BP} = 
\begin{cases} 
1 & \text{if } |c| > |r| \\ 
\exp\big(1 - \frac{|r|}{|c|}\big) & \text{if } |c| \le |r|
\end{cases},
\]
with $|c|$ the length of the candidate and $|r|$ the effective reference length.

\textbf{METEOR.\citep{banerjee2005meteor}}  
METEOR evaluates alignment at the unigram level, accounting for exact, stem, synonym, and paraphrase matches. For candidate $c$ and references  $\{r_k\}$:
\[
\text{METEOR} = F_{\text{mean}} \cdot (1 - \text{penalty}),
\]
where $F_{\text{mean}}$ is the harmonic mean of unigram precision and recall, and the penalty increases with the fragmentation of matches.

\textbf{CIDEr \citep{vedantam2015cider}.}  
CIDEr measures consensus with reference captions using TF-IDF weighting over N-grams:
\[
\text{CIDEr}(c, \{r_k\}) = \frac{1}{N} \sum_{n=1}^{N}\left( w_n \cdot \frac{1}{m} \sum_{k=1}^{m} \frac{\vec{g}_n(c) \cdot \vec{g}_n(r_k)}{\|\vec{g}_n(c)\| \, \|\vec{g}_n(r_k)\|}\right),
\]
where $\vec{g}_n(c)$ is the TF-IDF vector of N-grams in $c$, $w_n$ is typically uniform, and $N$ is the maximum N-gram length (commonly 4).

\textbf{BERTScore.\citep{zhang2020bertscore}}  
BERTScore evaluates semantic similarity at the embedding level. Let $E(c) = [e_1, \dots, e_{|c|}]$ and $E(r) = [r_1, \dots, r_{|r|}]$ be token embeddings of the candidate and reference captions, respectively. Then:
\[
\text{Precision: } P_B = \frac{1}{|c|} \sum_{i=1}^{|c|} \max_j e_i \cdot r_j, \quad
\text{Recall: } R_B = \frac{1}{|r|} \sum_{j=1}^{|r|} \max_i e_i \cdot r_j,
\]
\[
\text{F1: } F_B = \frac{2 \cdot P_B \cdot R_B}{P_B + R_B}.
\]

\textbf{SBERT Similarity \citep{reimers2019sentence}.}  
Sentence-BERT (SBERT) evaluates semantic similarity at the sentence embedding level rather than token-level alignment. 
Let $f(\cdot)$ denote the SBERT encoder that maps a caption into a fixed-dimensional embedding vector. 
Given a candidate caption $c$ and a reference caption $r$, their embeddings are:
\[
u = f(c), \quad v = f(r), \quad u, v \in \mathbb{R}^d.
\]

The similarity score is then computed using cosine similarity:
\[
\text{SBERTSim}(c, r) = \cos(u, v) = \frac{u \cdot v}{\|u\| \, \|v\|}.
\]

Here, $u \cdot v$ denotes the dot product, and $\|u\|$, $\|v\|$ are the Euclidean norms of the embeddings. 
The resulting score ranges from $-1$ to $1$, where higher values indicate greater semantic similarity between the candidate and reference captions.

\textbf{F1 Score.}
The F1 score, also known as the harmonic mean, provides a balanced measure of the model's performance by combining precision (the proportion of correctly predicted elements) and recall (the proportion of true elements captured). A higher F1 score indicates better overall alignment with reference captions.

\textbf{{Temporal Analysis \citep{sun2021nsp} , \citep{ilharco2019general}}.}
To evaluate discourse-level coherence, we leverage the Next Sentence Prediction (NSP) task from BERT-like models. 
Given two sentences $(s_i, s_j)$, the NSP classifier estimates the probability 
that $s_j$ logically follows $s_i$, denoted as $p_{\text{NSP}}(s_i, s_j) \in [0,1]$.
\paragraph{Align$_\text{DTW}$}
To assess temporal alignment between a predicted sequence and a reference sequence, we compute a Dynamic Time Warping (DTW) alignment over per-step embeddings.
Let $X = [x_1,\dots,x_T]$ and $Y = [y_1,\dots,y_S]$ denote embeddings (e.g., frame- or step-level), with a local cost
\[
d(x_i,y_j) \;=\; 1 - \cos(x_i,y_j) \;=\; 1 - \frac{x_i \cdot y_j}{\|x_i\|\,\|y_j\|}.
\]
Define the DTW dynamic program:
\[
D(i,j) \;=\; d(x_i,y_j) + \min\{ D(i-1,j), \; D(i,j-1), \; D(i-1,j-1) \},
\]
with $D(1,1)=d(x_1,y_1)$ and appropriate boundary conditions.
Let $\pi^\star=\{(i_k,j_k)\}_{k=1}^{K}$ be the optimal warping path obtained by backtracking from $(T,S)$, and $K=|\pi^\star|$.
We report an alignment \emph{similarity} by averaging cosine similarities along the optimal path:
\[
\text{Align$_\text{DTW}$}(X,Y) \;=\; \frac{1}{K}\sum_{(i,j)\in \pi^\star} \cos(x_i,y_j)
\;\in[-1,1].
\]
Higher values indicate better temporal alignment between the sequences.

\paragraph{Contradict$_\text{NLI}$}
To quantify logical inconsistency between a candidate caption and a reference, we apply a Natural Language Inference (NLI) model.
Let $g_{\text{NLI}}(c,r)$ produce a distribution over \{\textsc{entailment}, \textsc{neutral}, \textsc{contradiction}\}:
\[
\big(p_{\text{ent}}(c,r),\; p_{\text{neu}}(c,r),\; p_{\text{con}}(c,r)\big)
\;=\; g_{\text{NLI}}(c,r).
\]
The \emph{Contradict$_\text{NLI}$} score is the (average) probability of contradiction:
\[
\text{Contradict$_\text{NLI}$}(c,r) \;=\; p_{\text{con}}(c,r),
\qquad
\text{or}\qquad
\text{Contradict$_\text{NLI}$} \;=\; \frac{1}{N}\sum_{n=1}^{N} p_{\text{con}}(c_n,r_n)
\]
when aggregating over a set of pairs $\{(c_n,r_n)\}_{n=1}^{N}$.
Higher values indicate stronger evidence that the candidate contradicts the reference.

\paragraph{NSP=True.} 
For consecutive sentences in their original order, coherence is measured as:
\[
\text{NSP}_{\text{True}} = \frac{1}{N} \sum_{(s_i, s_{i+1}) \in D} p_{\text{NSP}}(s_i, s_{i+1}),
\]
where $D$ is the set of sentence pairs from the candidate text.

\paragraph{NSP=Shuffled.} 
For incoherent text constructed by randomly shuffling sentences, coherence is measured by:
\[
\text{NSP}_{\text{Shuffled}} = \frac{1}{N} \sum_{(s_i, s_j) \in D_{\text{shuf}}} p_{\text{NSP}}(s_i, s_j),
\]
where $D_{\text{shuf}}$ contains shuffled sentence pairs.

\paragraph{NSP=Delta.} 
To capture the relative coherence signal, we compute the difference between coherent and incoherent settings:
\[
\text{NSP}_{\Delta} = \text{NSP}_{\text{True}} - \text{NSP}_{\text{Shuffled}}.
\]

\noindent A higher $\text{NSP}_{\Delta}$ indicates that the model can better distinguish 
between coherent and incoherent sentence orderings.

\section{Temporal Metrics Analysis Tables}
\label{appendix: temporal metrics}
\begin{table}[ht]
\centering
\footnotesize
\setlength{\tabcolsep}{5pt} 

\begin{tabularx}{\textwidth}{>{\hsize=7\hsize\raggedright}X c c c c c}
\toprule
Method & Align$_\text{DTW}(\uparrow)$ & Contradict$_\text{NLI}(\downarrow)$ & 
\multicolumn{3}{c}{Coherence} \\
\cmidrule(lr){4-6}
 & & &  NSP$=$True($\uparrow$) & NSP$=$Shuffled($\downarrow$) & NSP$=$Delta($\uparrow$) \\
\midrule
GPT-4o & 2.86(0.14) & 0.03(0.00) & 1(0.00) & 1(0.00) & 0(0.00) \\
VLLaMA-3 & 3.15(0.02) & 0.06(0.00) & 0.99(0.00) & 0.99(0.00) & 0(0.00) \\
DynaStride & 3.01(0.00) & 0.05(0.00) & 1(0.00) & 0(0.00) & 1(0.00) \\
\bottomrule
\end{tabularx}
\vspace{0.1cm}
\caption{\footnotesize Alignment (DTW), contradiction (NLI), and Coherence metrics (true, shuffled, delta) on YouCook2 validation set for GPT-4o, VLLaMA-3, and DynaStride.}
\label{table:youcook2-dtw-contradict-coherence}
\end{table}

\begin{table}[ht]
\centering
\footnotesize
\setlength{\tabcolsep}{5pt}

\begin{tabularx}{\textwidth}{>{\hsize=7\hsize\raggedright}X c c c c c}
\toprule
Aggregator & Align$_\text{DTW}(\uparrow)$ & Contradict$_\text{NLI}(\downarrow)$ & 
\multicolumn{3}{c}{Coherence} \\
\cmidrule(lr){4-6}
 & & & NSP$=$True($\uparrow$) & NSP$=$Shuffled($\downarrow$) & NSP$=$Delta($\uparrow$) \\
\midrule
Phi & 3.14(0.03) & 0.03(0.00) & 1(0.00) & 1(0.00) & 0.00(0.00) \\
Mistral & 3.11(0.01) & 0.06(0.00) & 0.99(0.00) & 0.99(0.00) & 0.00(0.00) \\
Qwen3 & 3.01.00(0.00) & 0.05(0.00) & 1(0.00) & 0(0.00) & 1(0.00) \\
\bottomrule
\end{tabularx}
\caption{\footnotesize Alignment (DTW), contradiction (NLI), and Coherence metrics (true, shuffled, delta) for different aggregator models (Phi, Mistral, Qwen3).}
\label{table:aggregator-models-coherence}
\end{table}

\begin{table}[ht]
\centering
\footnotesize
\setlength{\tabcolsep}{5pt}

\begin{tabularx}{\textwidth}{>{\hsize=7\hsize\raggedright}X c c c c c}
\toprule
Model / & Align$_\text{DTW}(\uparrow)$ & Contradict$_\text{NLI}(\downarrow)$ & 
\multicolumn{3}{c}{Coherence} \\
\cmidrule(lr){4-6}
 Sample Freq.& & & NSP$=$True($\uparrow$) & NSP$=$Shuffled($\downarrow$) & NSP$=$Delta($\uparrow$) \\
\midrule
\textbf{GPT-4o} & & & & &  \\
\quad 5 Frames & 3.01(0.01) & 0.03(0.00) &1.00(0.00)& 1.00(0.00) & 0.00(0.00) \\
\quad 20 Frames & 2.89(0.00) & 0.03(0.00) & 1.00(0.00) & 1.00(0.00) & 0.00(0.00) \\
\quad 40 Frames & 2.91(0.00) & 0.02(0.00) & 1.00(0.00) & 0.99(0.00) & 0.00(0.00) \\
\textbf{VLLaMA-3} & & & &  & \\
\quad 5 Frames & 3.21(0.00) & 0.05(0.00) & 0.99(0.00) & 0.99(0.00) & 0.00(0.00) \\
\quad 20 Frames & 3.11(0.02) & 0.07(0.01) & 0.99(0.00) & 0.99(0.00) & 0.00(0.00) \\
\quad 40 Frames & 3.11(0.01) & 0.08(0.01) & 0.99(0.00) & 0.99(0.00) & 0.00(0.00) \\
\textbf{DynaStride} & & & &  & \\
\quad 5 Frames & 0.99(0.00) & 0.06(0.00) & 0.99(0.00) & 0.99(0.00) & 0.00(0.00) \\
\quad 20 Frames & 2.96(0.00) & 0.04(0.00) & 0.99(0.00) & 0.99(0.00) & 0.00(0.00) \\
\quad 40 Frames & 2.87(0.00) & 0.05(0.00) & 1.00(0.00) & 1.00(0.00) & 0.00(0.00) \\
\bottomrule
\end{tabularx}
\caption{\footnotesize Alignment (DTW), contradiction (NLI), and Coherence\_NSP metrics (true, shuffled, delta) for different frame sampling rates (every 5, 20, and 40 frames).}
\label{table:frame-sampling-coherence}
\end{table}

\section{Scene Level Caption Comparisons}
\label{appendix: caption comparison}
\subsection{Aggregator Models vs Baseline Models}
\includegraphics[width=\linewidth]{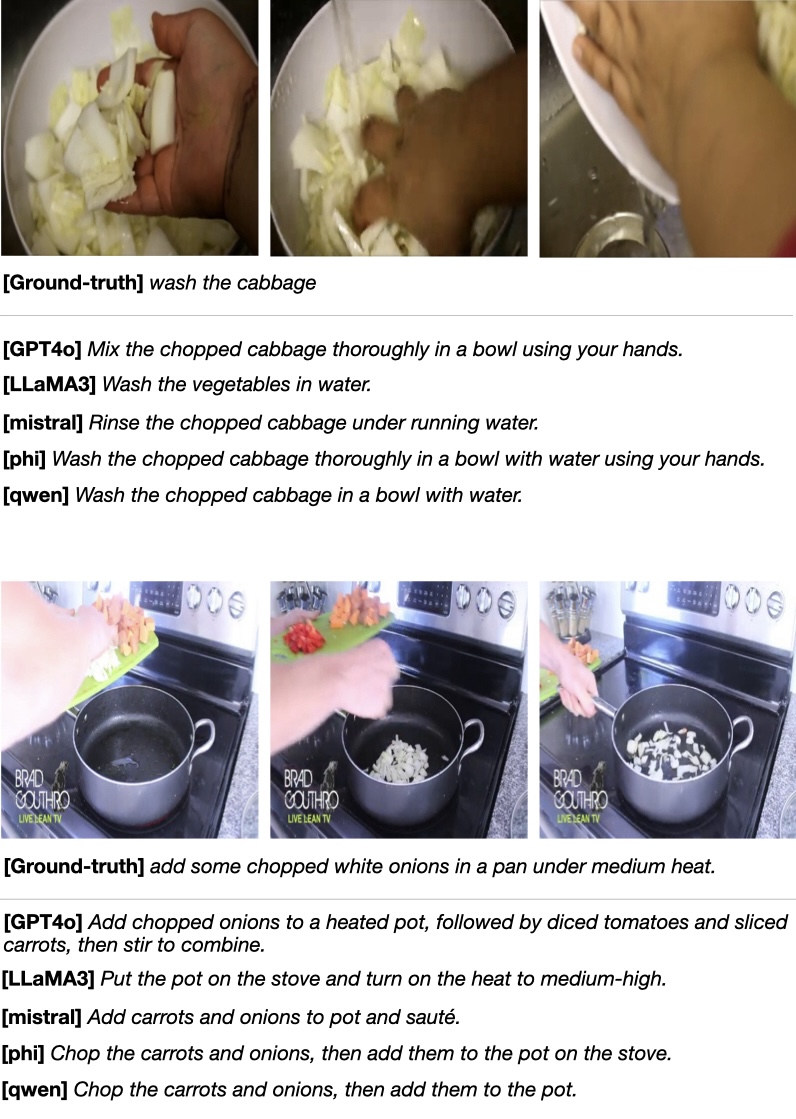}
\newpage
\subsection{Frame Sampling Rate Caption Comparisons}
\includegraphics[width=\linewidth, height=18cm]{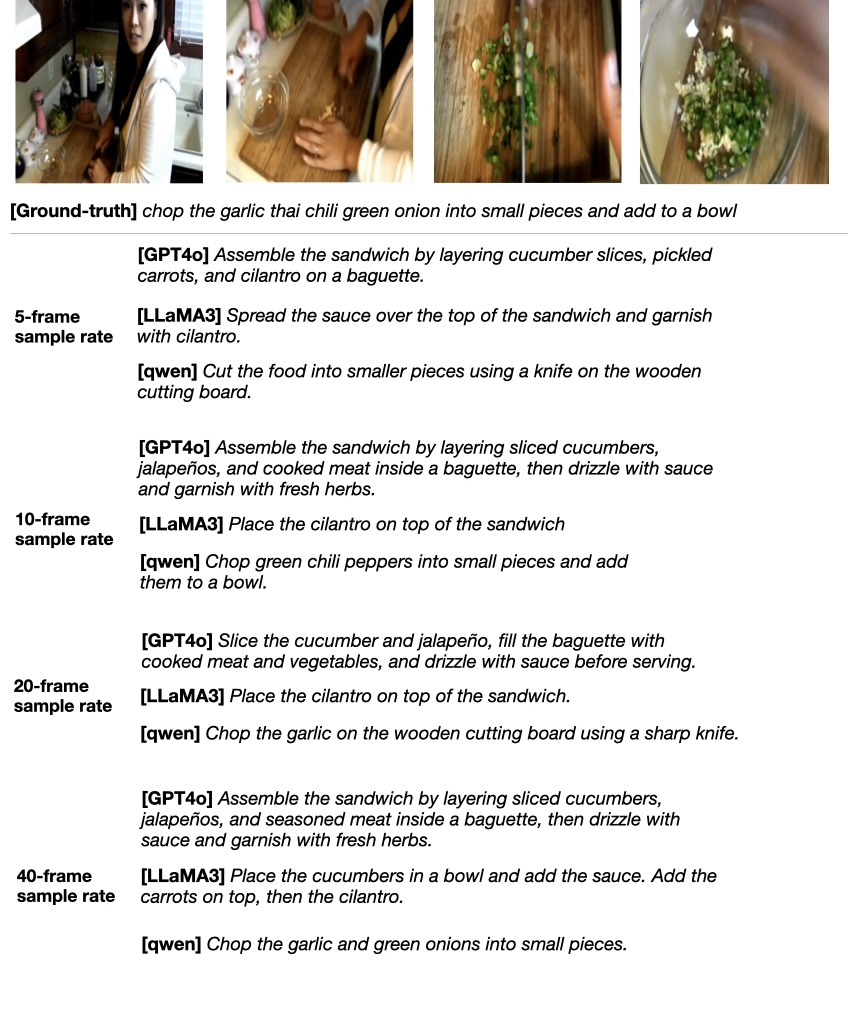}

\newpage

\section{Plots of Frame Sampling Rate vs Metrics}
\label{appendix: frame sampling vs metrics}
\subsection{N-gram Metrics}

\includegraphics[width=\textwidth]{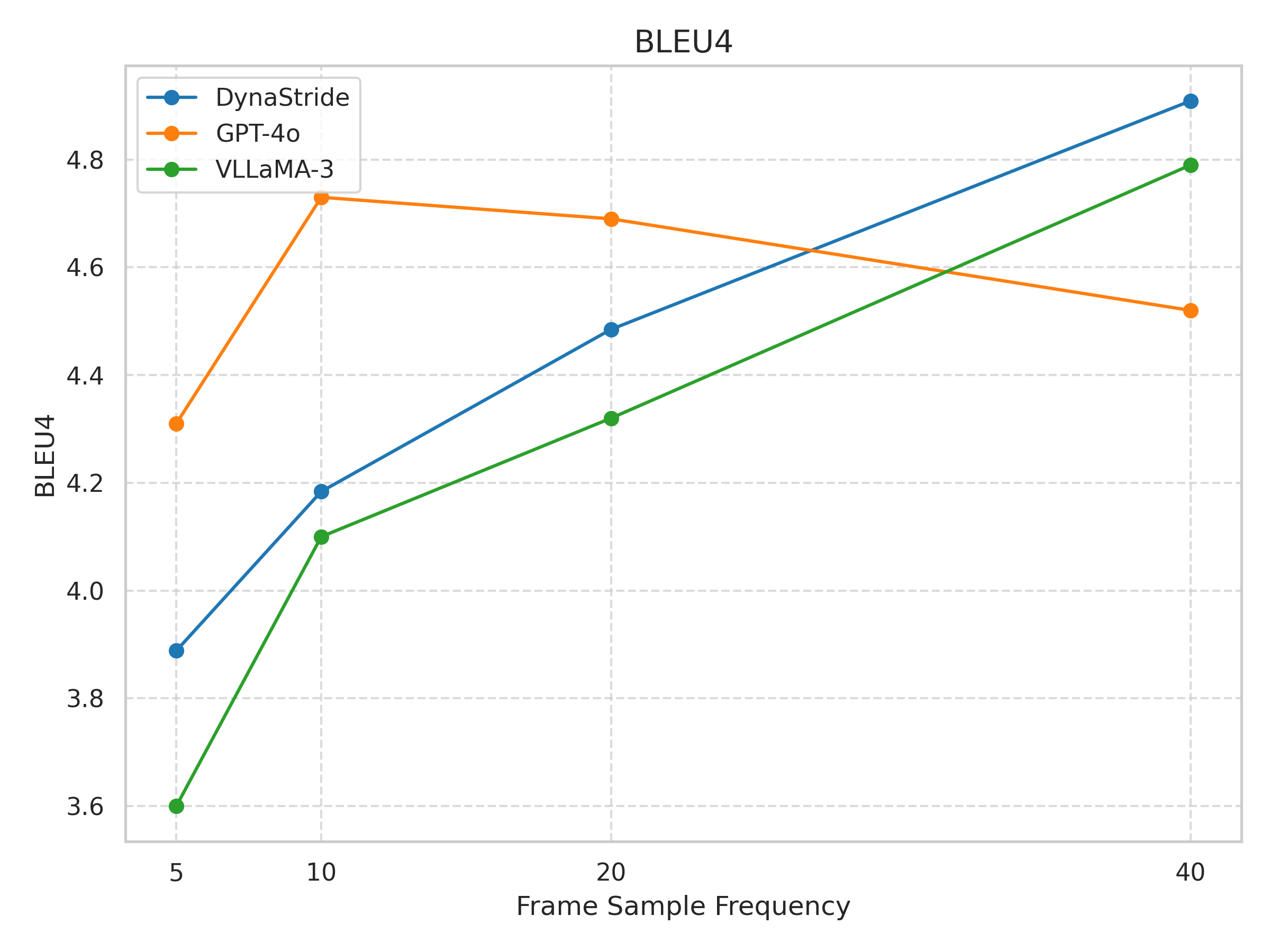}

\includegraphics[width=\textwidth]{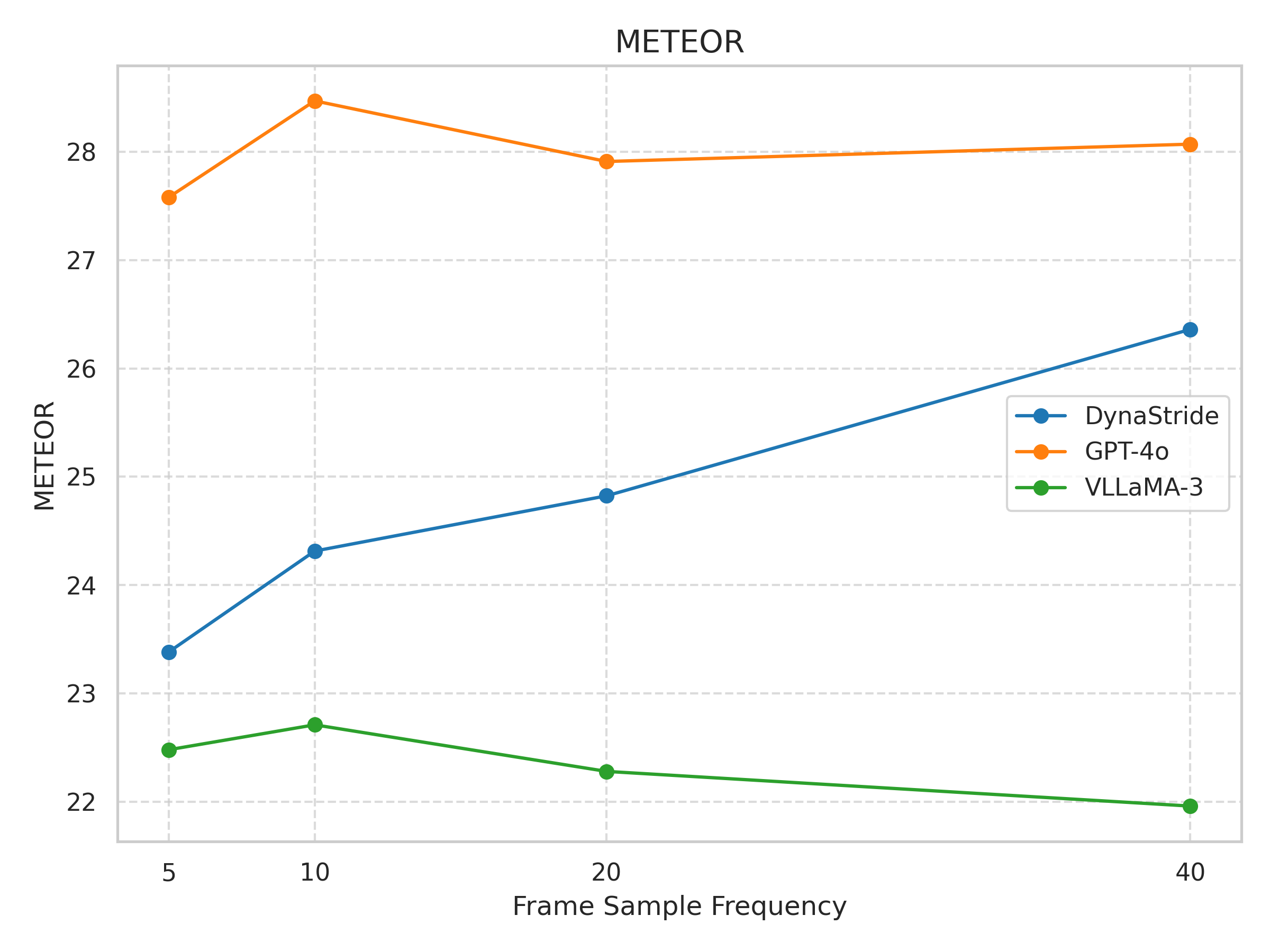}

\includegraphics[width=\textwidth]{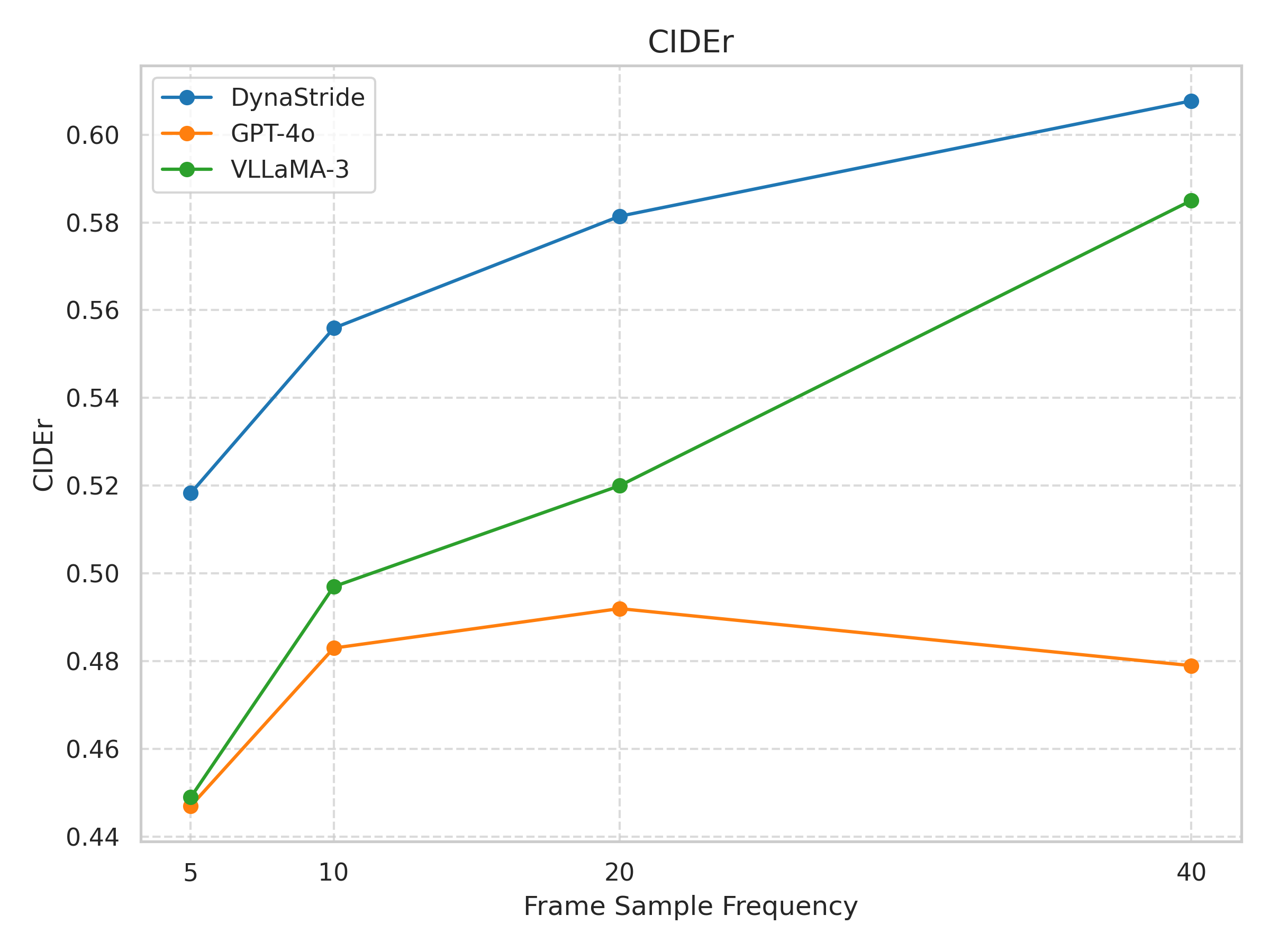}

\subsection{Semantic Metrics}

\includegraphics[width=\textwidth]{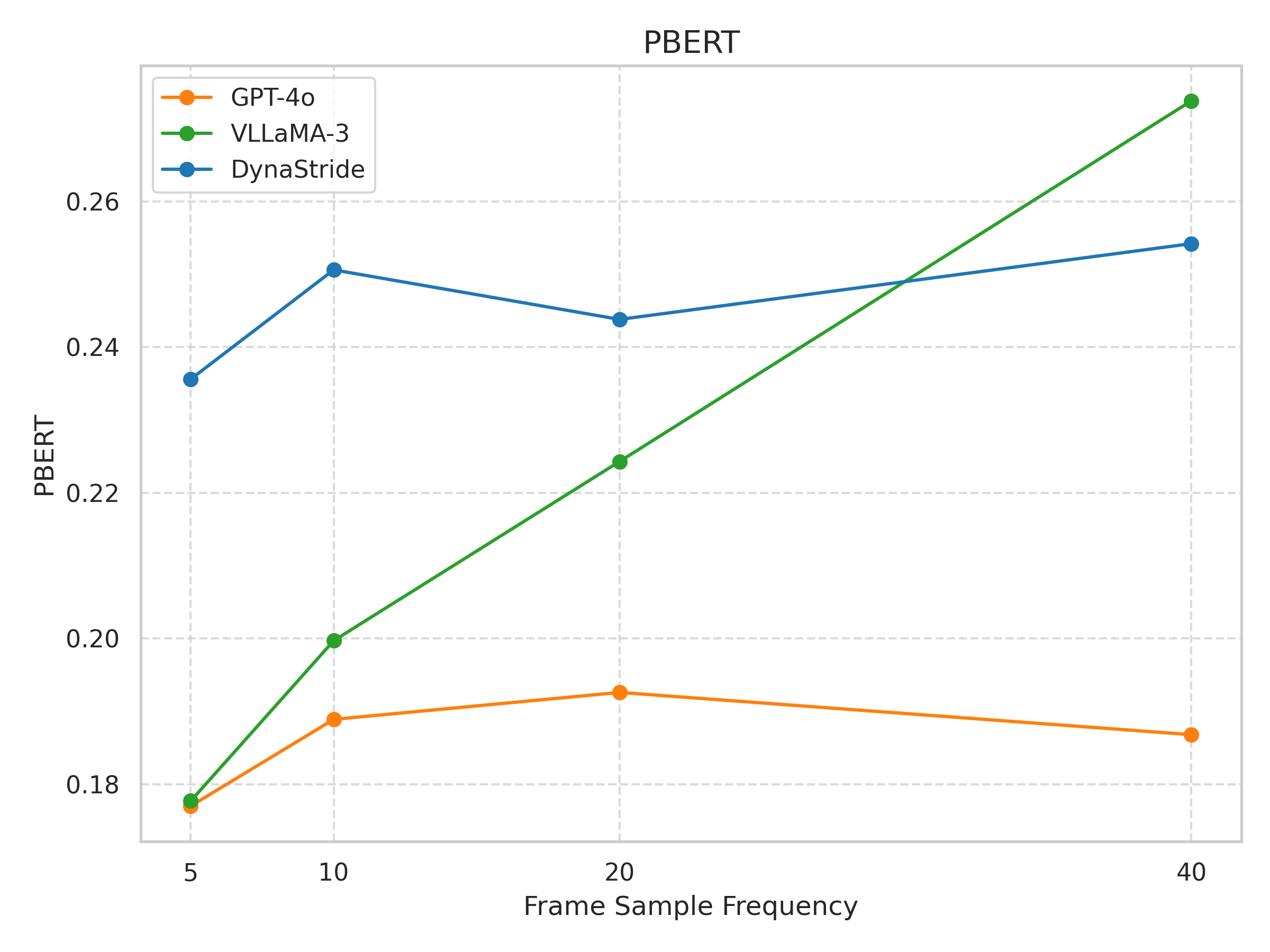}

\includegraphics[width=\textwidth]{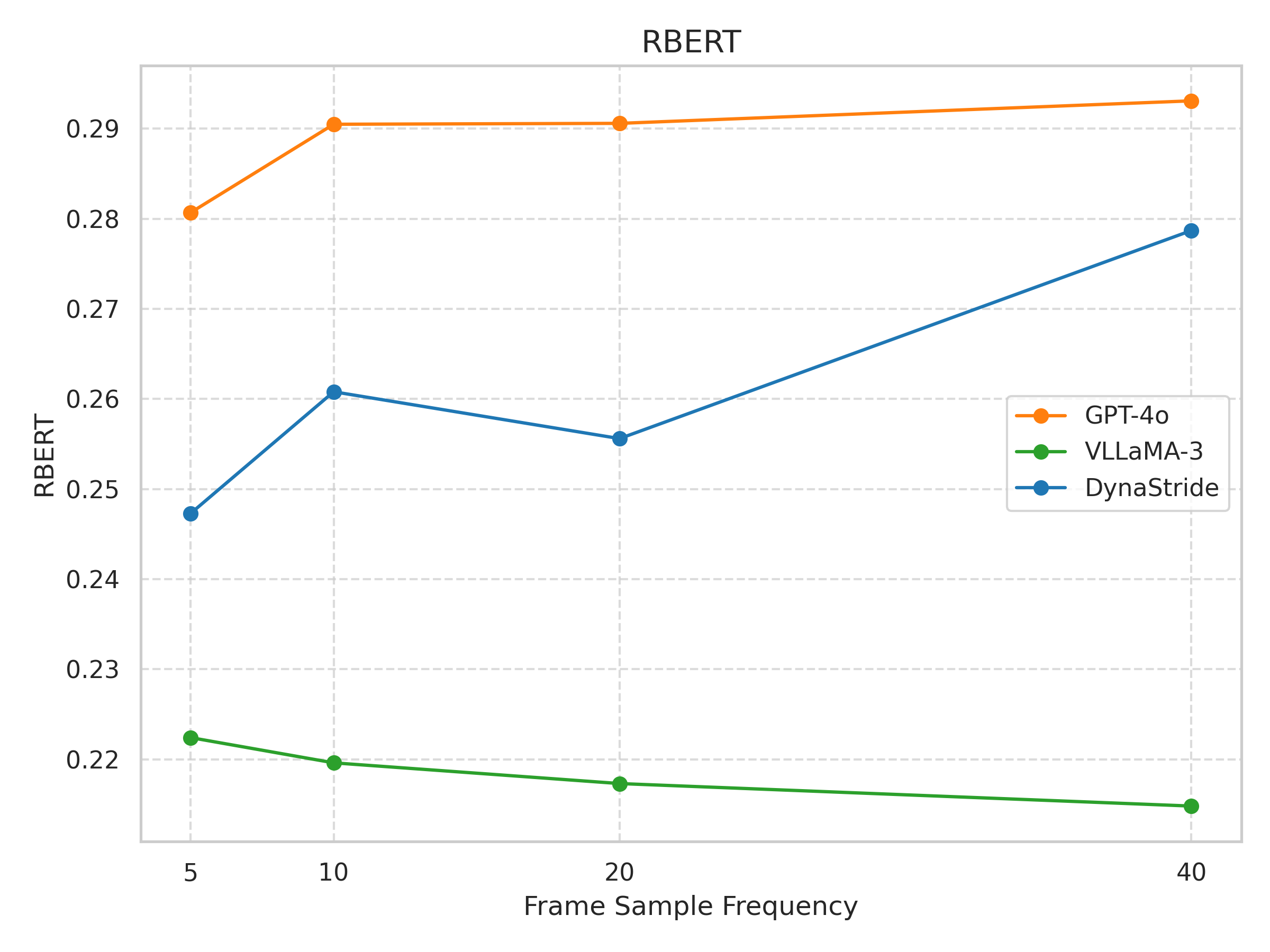}

\includegraphics[width=\textwidth]{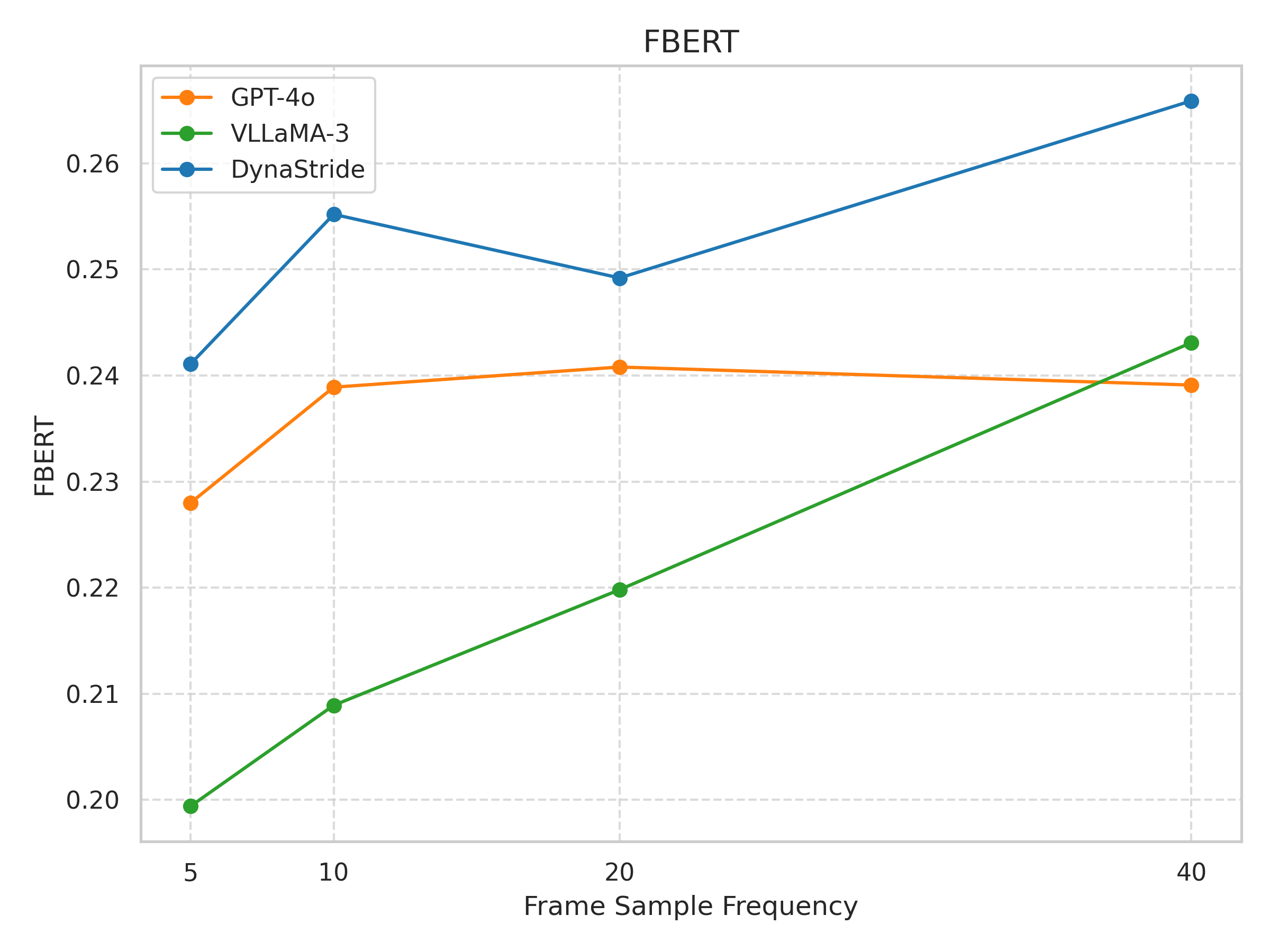}

\includegraphics[width=\textwidth]{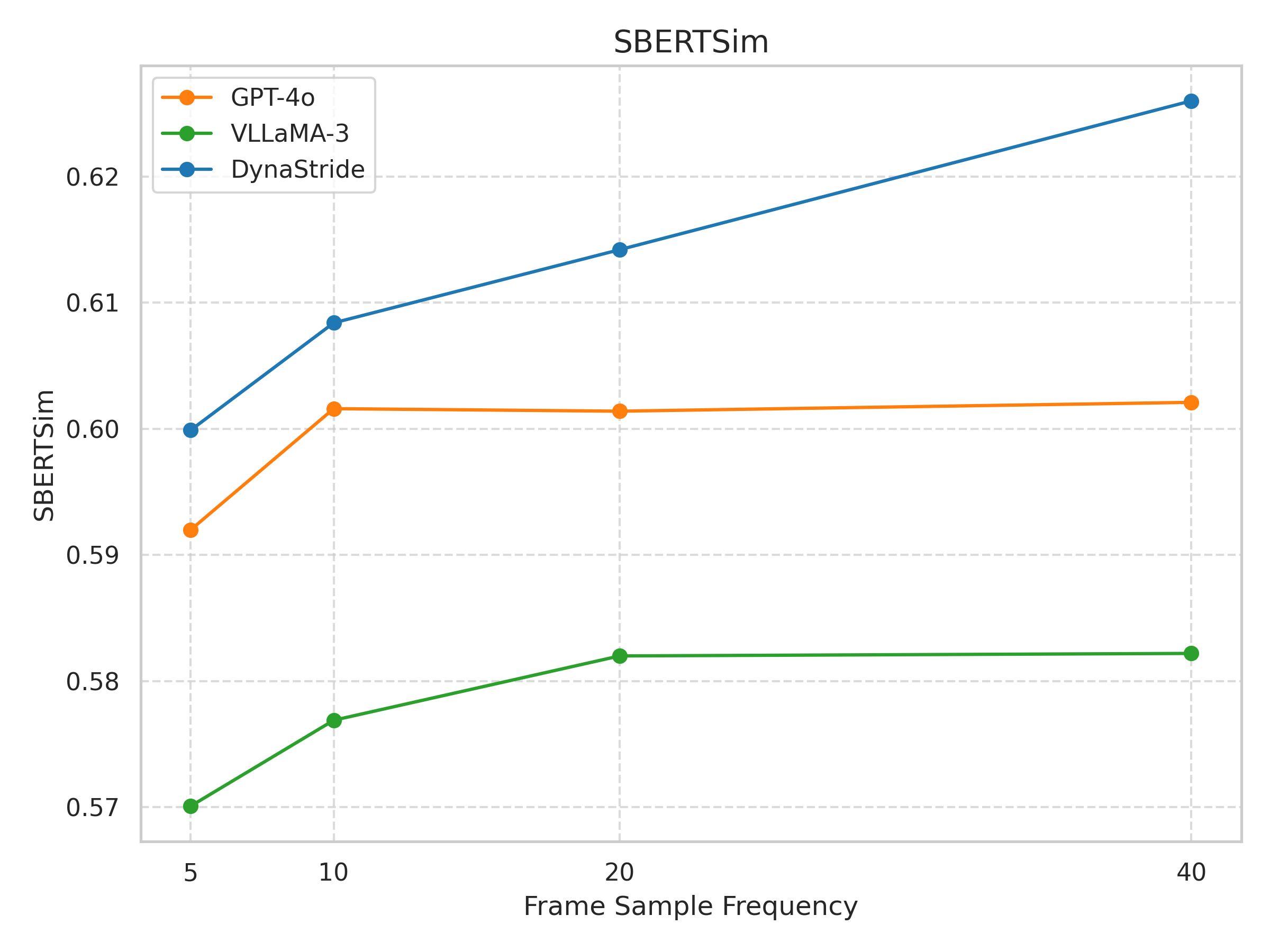}

\subsection{Temporal Metrics}

\includegraphics[width=\textwidth]{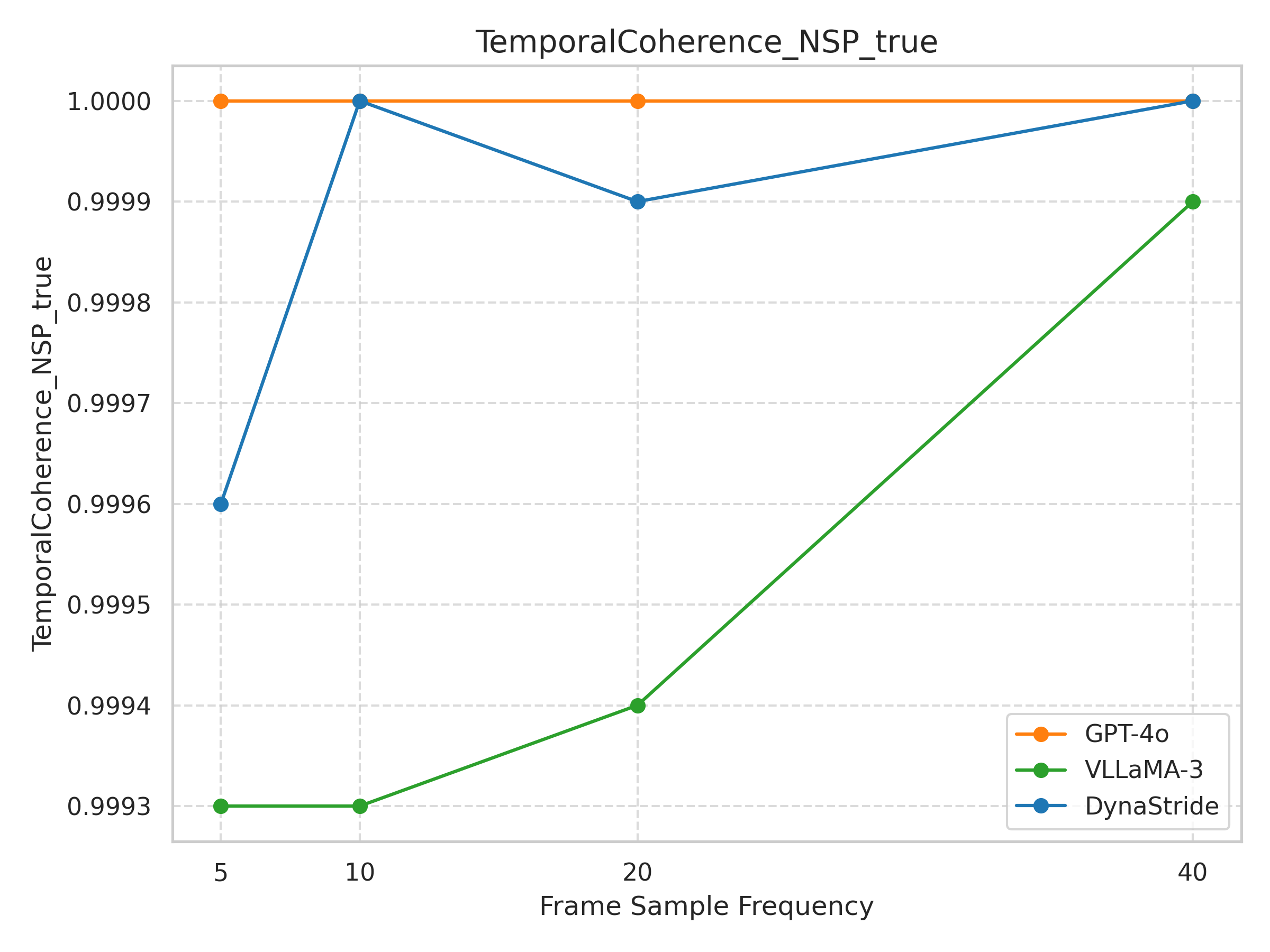}

\includegraphics[width=\textwidth]{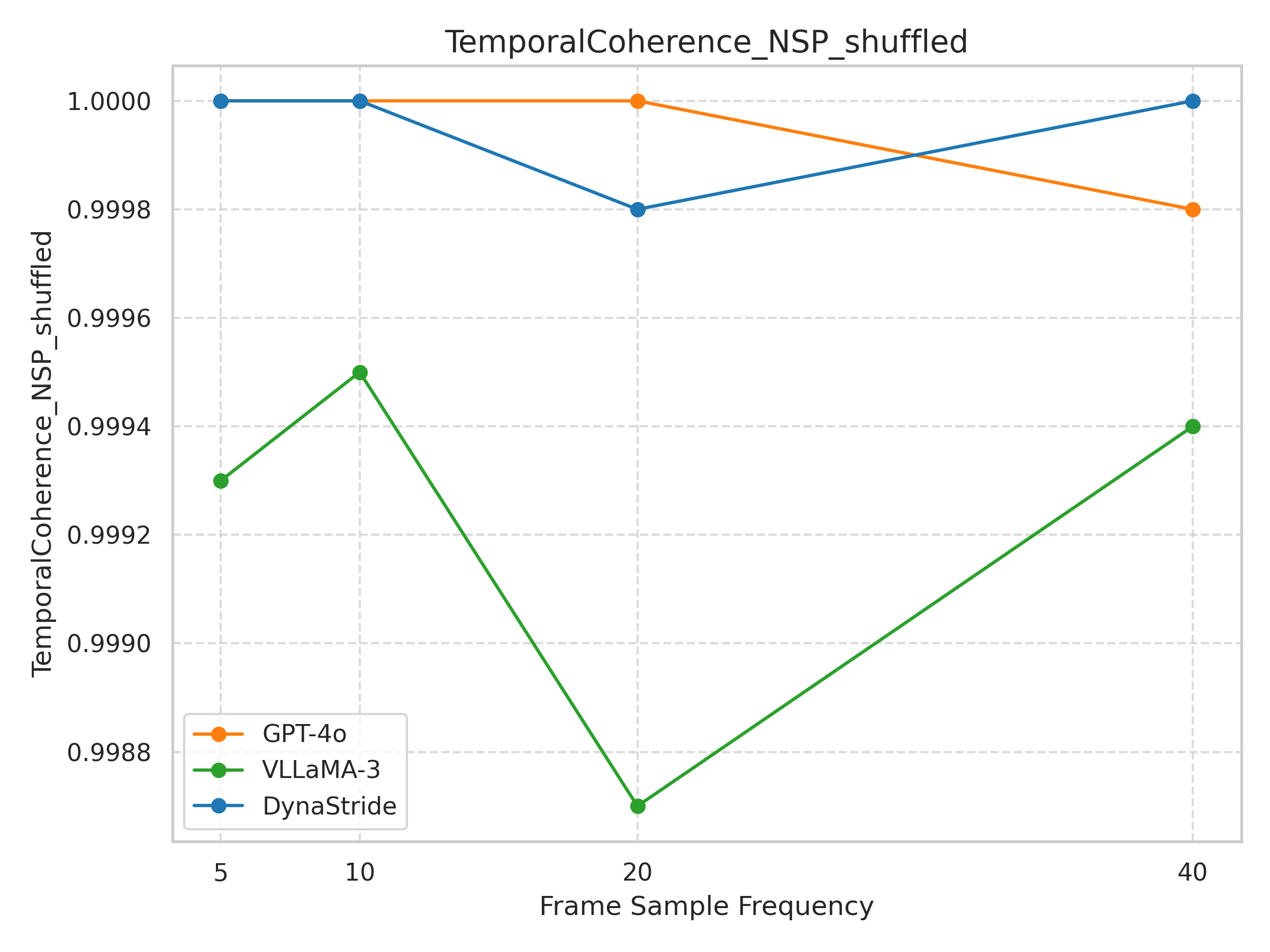}

\includegraphics[width=\textwidth]{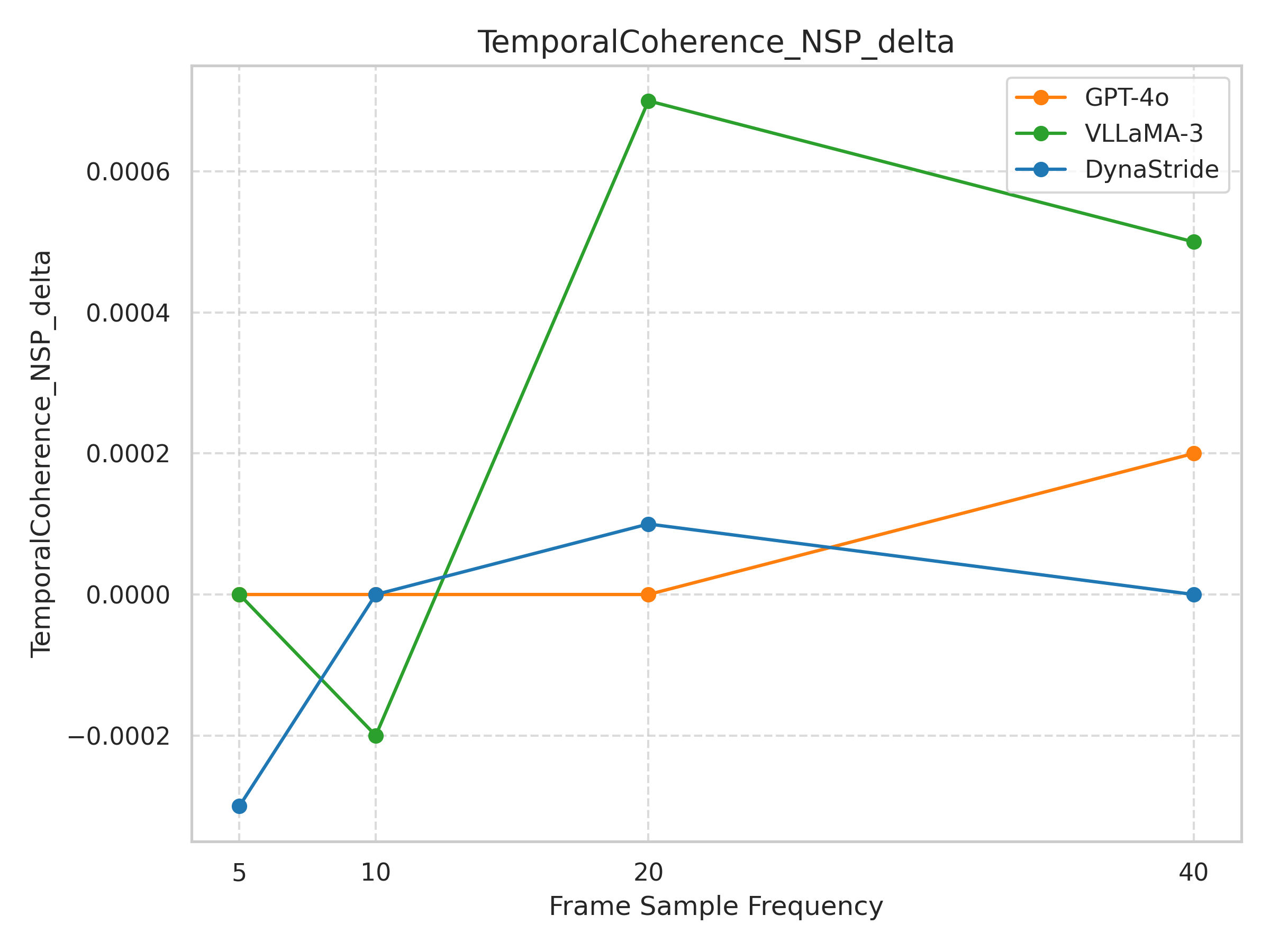}

\includegraphics[width=\textwidth]{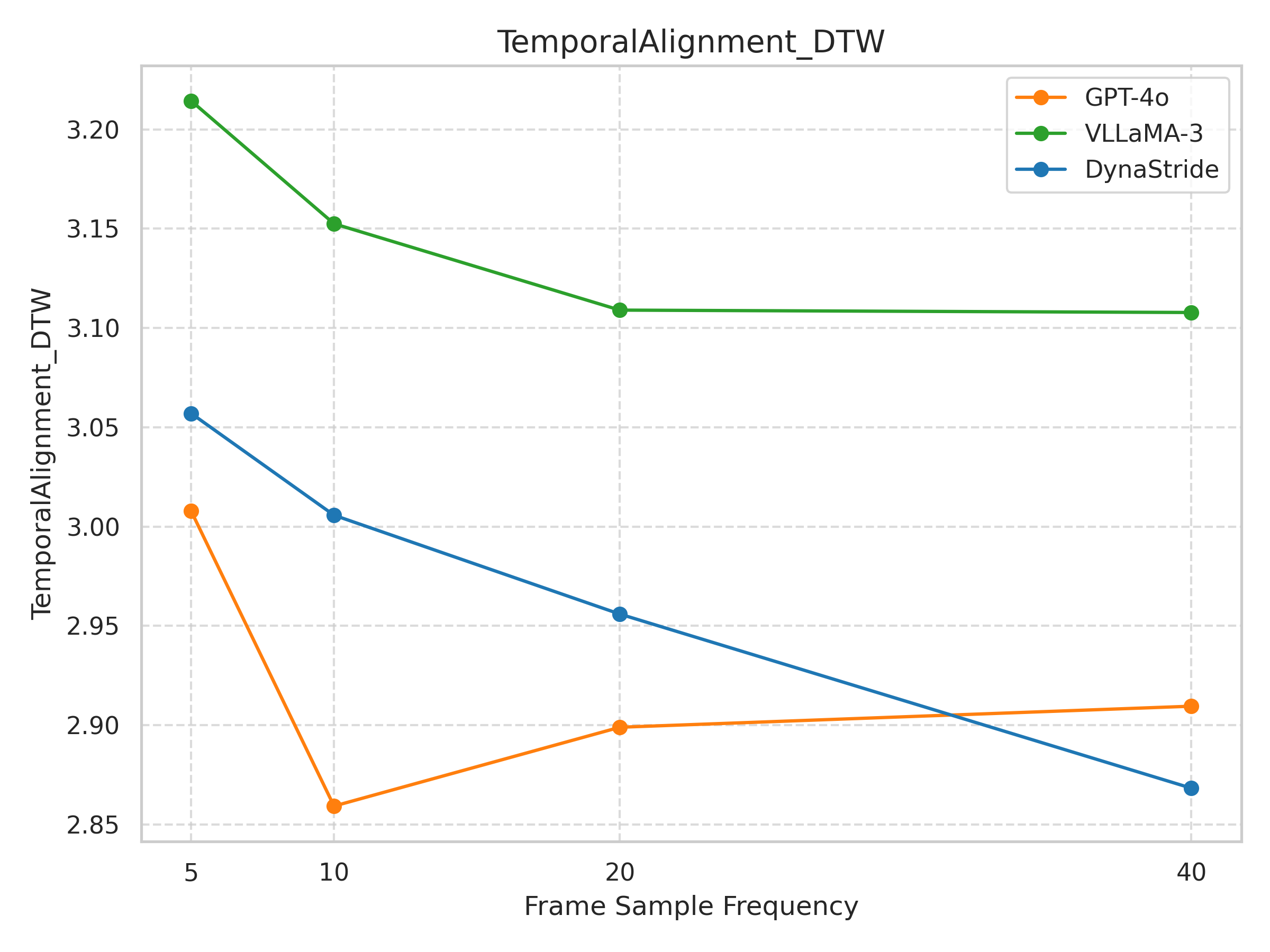}

\includegraphics[width=\textwidth]{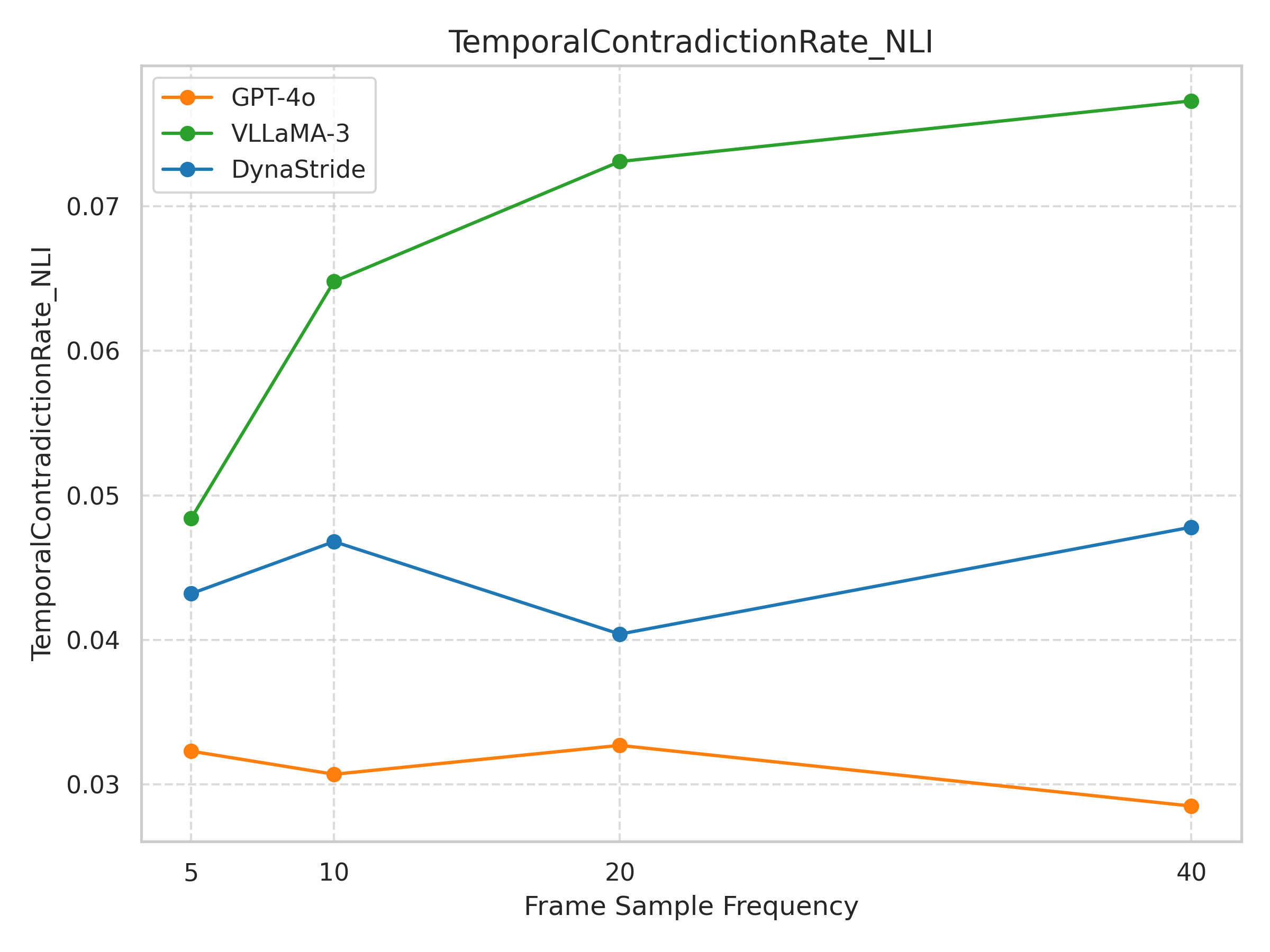}

\end{document}